\documentclass[lettersize,journal]{IEEEtran}
\usepackage{amsmath,amsfonts}
\usepackage{amssymb}
\usepackage{algorithmic}
\usepackage{algorithm}
\usepackage{array}
\usepackage[caption=false,font=normalsize,labelfont=sf,textfont=sf]{subfig}
\usepackage{textcomp}
\usepackage{stfloats}
\usepackage{url}
\usepackage{verbatim}
\usepackage{graphicx}
\usepackage{cite}  
\usepackage{dsfont}
\usepackage{booktabs}

\usepackage{bm}
\usepackage{color} 
\usepackage{multirow}

\begin{document} 
	

\title{{Multi-robot Social-aware Cooperative Planning in Pedestrian Environments Using Multi-agent Reinforcement Learning}

\author{Zichen He, Chunwei Song, Lu Dong*}

\thanks{First Version of Manuscript. Submitted on 2022-11-29 }}

\markboth{IEEEXplore}
{Shell \MakeLowercase{\textit{et al.}}: A Sample Article Using IEEEtran.cls for IEEE Journals}


\maketitle

\begin{abstract}
Safe and efficient co-planning of multiple robots in pedestrian participation environments is promising for applications.  In this work, a novel multi-robot social-aware efficient cooperative planner that on the basis of off-policy multi-agent reinforcement learning (MARL) under partial dimension-varying observation and imperfect perception conditions is proposed.  We adopt temporal-spatial graph (TSG)-based social encoder to better extract the importance  of social relation between each robot and the pedestrians in its field of view (FOV). Also, we introduce K-step lookahead reward setting in multi-robot RL framework to avoid aggressive, intrusive, short-sighted, and unnatural motion decisions generated by robots.  Moreover, we improve the traditional centralized critic network with multi-head global attention module to better aggregates local observation information among different robots to guide the process of  individual policy update. Finally, multi-group experimental results verify the effectiveness of the proposed cooperative motion planner. 
\end{abstract}


\section{Introduction}

\IEEEPARstart{W}{ith} the development of robotics and artificial intelligence, autonomous mobile robots are gradually deployed in the daily life of people.  For example, in scenarios such as airports, campuses, unmanned supermarkets, and smart warehouses, service mobile robots cannot avoid interacting with multiple pedestrians \cite{he2021review}.  Therefore, it is a very significant research hotspot to teach robots with social safety awareness.  Moreover,  in the application scenarios described above, the single robot often faces problems such as limited sensing range,  low planning efficiency, and weak stability during the operation.  In contrast, Multi-robot cooperation can better share the unknown information of the environment from multiple perspectives between individuals.  This facilitates the perception ability of each robot and ultimately improves the planning efficiency. 

\begin{figure}[!t]
	\centering
	\includegraphics[width=3.1 in]{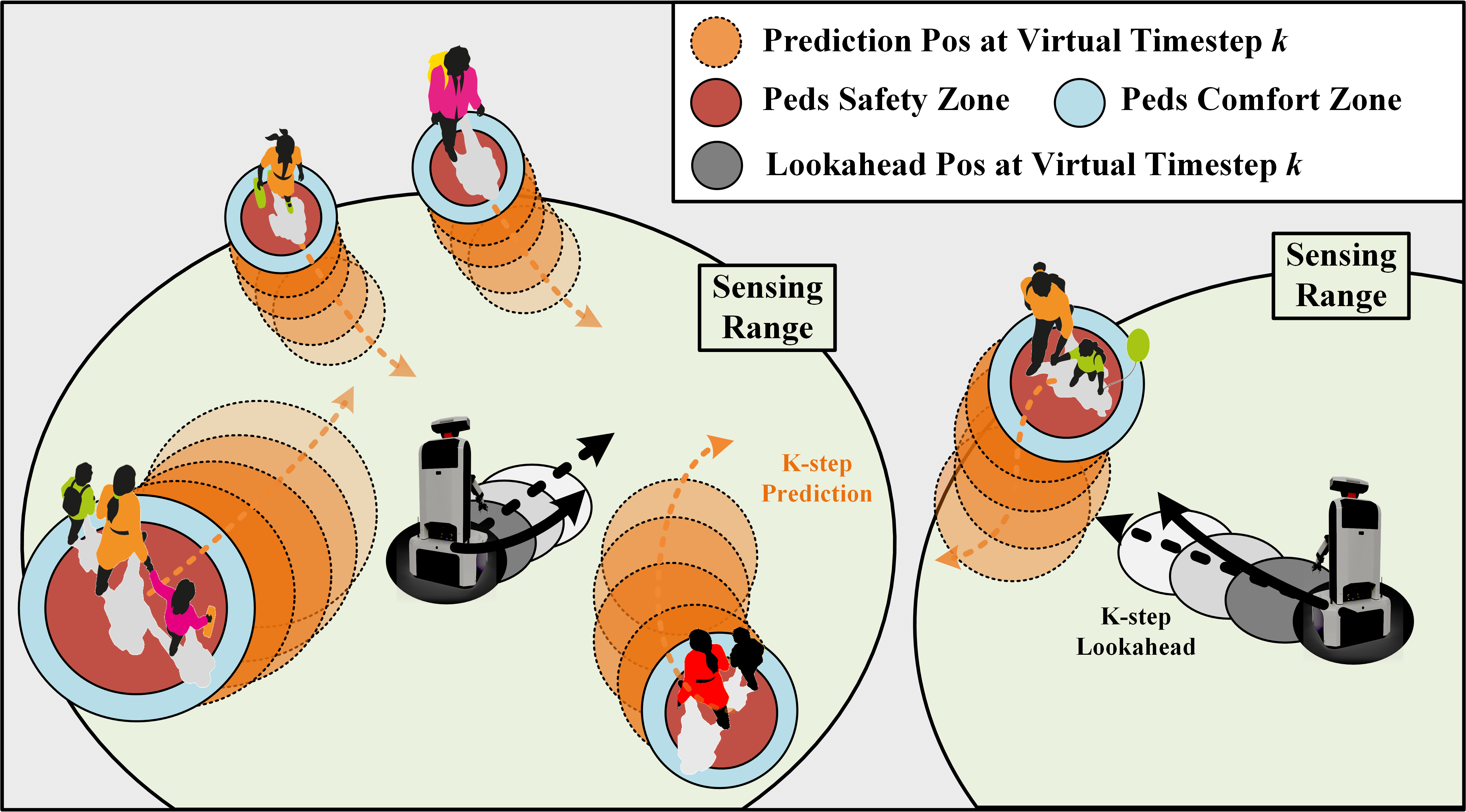}
	\caption{Multiple robots with limited sensing range perform decentralized cooperative inference planning in the pedestrian participation environment. Social-aware and CTDE-based multi-agent reinforcement learning architecture helps robots to interact better with pedestrians while planning efficiently and collaboratively. Pedestrians prediction module-based K-step lookahead interaction reward item motivates the robot to generate stronger awareness of crowd safety.}
	\label{Fig1_Content} 
\end{figure} 

Unlike pure robots co-planning scenarios in \cite{9773295,9953939}, communication between humans and robots is not possible in social interaction process.  Thus, it is highly challenging for each robot to perform autonomous collaborative navigation in pedestrian-rich environments.  Early RL-based works \cite{phillips2011sipp,9205217,9354210} tend to treat pedestrians as dynamic obstacles with simple state-update kinematics. This setting is convenient for robots to handle, but the behavior of pedestrians in reality has certain social properties and uncertainties. For example, pedestrians may suddenly change their motion speed, or change their target points.  Robots unable to generate proper policies to cope with such situations might cause unsafe issues. In \cite{8593871,9317723,9001226,matsuzaki2022learning}, researchers design specialized one-step comfort zone intrusion penalty function to improve the social safety of human-robot interaction.  Although this setting is more practical, they choose to deal with the relative social relationship between robots and humans on the basis of simple proximity function. In real-world scenarios,  some pedestrians near the robot may be moving in the same direction or away from it.  The attentional importance of these pedestrians may not be as critical as others further away but moving towards this robot.  Since these humans are more likely to collide with the current robot.  Meanwhile, one-step reward consideration only would make the robot short-sighted \cite{8794134}. In \cite{zhou2021r, 9341519}, researchers assume that each robot is fully perceptive of environments.  This setting is suitable for handling small scene tasks. In realistic deployment, the field of view (FOV) of every robot is limited by the sensing range of the dominant sensor. Hence the robot often lacks global information and has to tackle the variable observation dimensions of the human flow within the FOV \cite{9561595}. 

At the level of multi-robot cooperative operations,  centralized works like \cite{tang2018hold,machines9040077,yu2021surprising} suffer from scalability and the lack of effective global information aggregation patterns.  Decentralized approaches like \cite{5980392,8661608,liu2022socially} are efficient but prone to local optimality and self-interested issues.  Our previous work \cite{9953939} based on centralized training and decentralized execution (CTDE) MARL paradigm and has been demonstrated to achieve state-of-the-art results in dense and pure-robot co-planning tasks.  We want to extend this architecture to further address pedestrian participation multi-robot co-planning.  

To address problems mentioned above, we propose a novel social-aware multi-robot co-planning method called Multi-agent Social-Aware Attention-based Actor-Critic (MSA3C) for the task scenario described in Fig. \ref{Fig1_Content}.  First, we introduce the attention-based centralized critic to better aggregate multiple observations from different perspectives of multiple robots, and we utilize its output value to provide better guidance to each robot for local policy updates.  Also, we design a rollout data processing pipeline to adapt to off-policy MARL setting and handle the variable dimension issue of human flow observations within the FOV of each robot. This trick is much more practical and can effectively improve the scalability of our method. Then, we adopt temporal-spatial graph (TSG)-based social encoder to extract the relative importance of the surrounding pedestrians to assist each robot make better social decisions.  In addition,  to enhance the social comfort awareness of each robot and avoid unnatural, aggressive, and short-sighted decisions,  we introduce K-step lookahead reward function during the training phase to better evaluate the impact of current actions of robots on future human-robot interaction. To sum up, our main contributions are as follows:
\begin{itemize}
\item We propose a CTDE-based MSA3C framework for handling multi-robot cooperative planning with social safety and comfort awareness under limited FOV condition. Our method achieves great performance in multiple experiments compared in comparison to various baselines.
\item We design a multi-agent rollout replaybuffer to cope with variable dimension of transitions and introduce parameter sharing social encoder for each robot on the basis of TSG to help robot better understand relative social relationship of surrounding pedestrians.
\item We integrate K-step lookahead reward function in MARL paradigm during the training phase to enhance the social comfort awareness of each robot and avoid unnatural and shortsighted policies.
\end{itemize}

\begin{figure*}[]
	\centering
	\includegraphics[width=6.3 in]{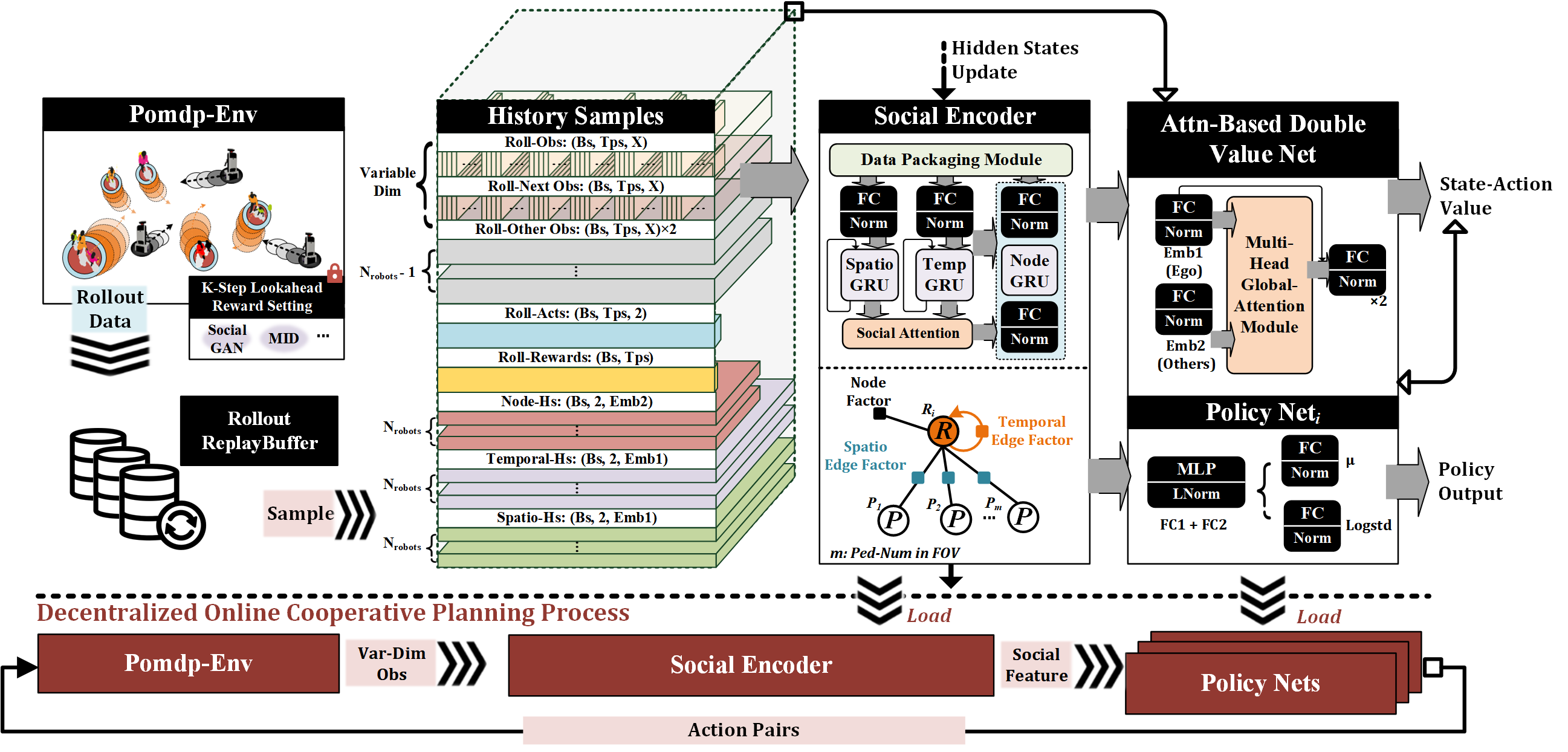}
	\caption{The overall architecture of Multi-robot Social-Aware Attention-based Actor-Critic (MSA3C) algorithm. }
	\label{Fig2_Tot_Alg_Arch} 
\end{figure*}  

\section{Related Work}
\subsection{Multi-robot Cooperative Planning} 
Classical multi-robot co-planning methods can be divided into centralized methods and decentralized methods.  Among centralized methods, \cite{6225009} is a typical optimization-based collaborative trajectory generation method. \cite{7539623} is a typical co-planning work on the basis of heuristic search. Centralized methods has completeness or probability completeness. However, these methods rely on accurate global information acquisition or aggregation approaches. Also, they are suffers from several issues such as low scalability and discrete dimensional explosion.  As for the decentralized pattern, Desaraju et al. in \cite{5980392} present decentralized multi-agent rapidly exploring random tree and can sample multiple feasible paths for multiple robots simultaneously.  The application of this approach requires the construction of explicit communication channels between robots.  Reaction-based velocity obstacle (VO) methods \cite{8516848} have been widely researched for their high efficiency and real-time property.  Reciprocal VO (RVO) \cite{5746538} improves the oscillation issue. In \cite{berg2011reciprocal},  optimal reciprocal collision avoidance (ORCA) and its variants are presented to further improve the optimality of RVO.  In \cite{8581239,8834353}, researchers extend the application range of ORCA to handle heterogeneous and non-holonomic constraints existing co-planning tasks.  VO-based approaches rely on perfect perception conditions and suffer from freezing robot issue. Moreover, such method cannot be generalized well across different scenarios without the cumbersome hand-crafted process.

\subsection{Learning-based Social Aware Robot Motion Planning}
Currently, mobile robots have been widely deployed in real-life scenarios where humans are involved \cite{he2021review}.  Therefore, relevant studies related to teaching robots to learn to interact reasonably with pedestrians and decrease the intrusion into their motion comfort zones have attracted  the attention of many researchers \cite{8794134}.  Meanwhile, the development of deep learning (DL) and RL has provided novel avenues for data-driven social aware planning technologies.  Michael Everett et al. have been working on promoting the research on decentralized multi-robot motion planning deployed in pedestrian environments \cite{8593871,9317723,9001226}.  Their research works have also inspired later researchers \cite{fan2020distributed, 8794134, 9561595, matsuzaki2022learning}. However, these works only integrate one-step human-robot interaction reward and handle social relation on the basis of proximity function. The robot cannot fully understand the importance of each pedestrian in its FOV and is prone to be short-sighted.  Some researcher select raw sensor data (e.g., 2D laserscans \cite{qiu2022learning} ) as input to design end-to-end multi-agent RL pattern to handle multi-robot co planning in real scenes.  The sensor-level approaches overcome the problem of variable observation dimension but introduce large-size input and interpretability issues.  Moreover, the MARL framework used by these methods adopts a simple concatenation trick \cite{yu2021surprising} to aggregate joint observations from different robots. This unfocused way brings unnecessary redundant information to individuals. In \cite{9154607,9091314}, scholars introduce supervised learning to teach the robot planning policy. The real performance of these methods rely on the quality of labelled data or demonstrations.

\section{Methodology}  
\subsection{Dec-POMDP Configuration}
First, we model the problem of multi-robot cooperative planning in human crows as a decentralized partially observable Markov decision problem (Dec-POMDP) consisting of the tuple $G=(\mathcal{N},S,O,A,P,r, \Omega)$. Here, $\mathcal{N} =\{1,...,n \}$ is a finite set describing the number of robots in the environment. $s\in S$ represents the full state space of all agents (including robots and pedestrians). $o_i \in \Omega \sim O(s,i) $ denotes the partial observation of each robot $i$ for the external world at each time-step. $O$ is the observation kernel. $P(s^\prime|s,u)$ is the state transition function of the environment. $A$ represents the joint action space of all robots. $r$ is the team reward for all robots. 

\subsubsection{Observation} 
In this paper, the observation setting of the robot $i$ at time-step $t$ is as follows: 
\begin{equation}
	\textbf{o}_t = [ \textbf{s}_{\text{ego}}, \textbf{o}_{\text{other} }]
\end{equation} 
where $\textbf{s}_{\text{ego}} = [p_x,p_y,r,g_x,g_y,v_{\text{pref}}, v_x, v_y, \theta] $ is the full state vector of the ego-robot, including the position, the radius of the safety domain, the relative position of the target point, the preferred velocity and the orientation. $\textbf{o}_{\text{other}}$ is an imperfect perception vector with variable dimensionality. It only integrates the relative position of all agents (including pedestrians and other robots) in the FOV of ego-robot at time-step $t$. 

\subsubsection{Action Space} 
All robots in this paper are holonomic. This setting increases the training efficiency. We can utilize the hybrid planning framework in our previous work \cite{9773295} to handle extra optimization objectives and constraints, and improve the quality of the final motion trajectory at the back-end. In this paper, each robot have continuous action space $a_t = [v_x,v_y]$ in the range of $[v_{\text{min}}, v_{\text{max}}]$. These two parameters depend on the performance of motors. In fact, we scale the speed range to $[-1,1]$ during the training phase and rescale it during the environment update phase. 

\subsubsection{Reward Setting}
In this paper, we split the reward setting of the robot $i$ into a basic configuration and a K-step lookahead reward setting. Note that we assume that each robot has integrated object detection module and can accurately identify whether the interaction object is a human or another robot.

The total reward configuration is shown as follows: 
\begin{equation}\label{eq_basic-reward}
	r_{R_i}=\left\{\begin{array}{ll}
		-1 & \text { if } \text{any}(\textbf{d}_{\text{RRs}})<0\text{ or } \\ 
		& \text{any}(\textbf{d}_{\text{RPs}})<0  \\  
		-F_{\text{scale}}|d_{g}^t| + r_{\text{time}}   & \text {otherwise}. \\ 
		+ r_{\text{comfort}} + r_{K_{\text{step}}}
	\end{array}\right.    
\end{equation} 
where $\textbf{d}_{\text{RRs}}=[d_{R_i R_j}-r_{R_i}-r_{R_j},...]$ represents the relative distance between the robot $i$ and all other robots in its FOV. $r_{R_i}$ and $r_{R_j}$ represent the radius of the safety zone of the robot $i$ and the robot $j$. $\textbf{d}_{\text{RPs}} = [d_{R_i P_j}-r_{R_i}-r_{P_j},...]$ is the relative social distance vector between the robot $i$ and other pedestrians in its FOV. $r_{P_j}$ denotes the safety zone of the pedestrian $j$. $d_{g}^t = ||p_g^t - p^t_r||_2$ is the relative L2 distance between the robot $i$ and the target point at time-step $t$. $F_{\text{scale}} = {\frac{1}{R^2_\text{env}}}$ is the scaling constants. $R_{\text{env}}$ is the radius of the current scenario. $r_{\text{time}} = -0.001$ is the time penalty. This items promotes the robot to reach the goal as quickly as possible. Also, $r_{\text{comfort}} = - \mathds{1} \cdot  0.5$. $\mathds{1}$ indicates whether the robot $i$ invades the social comfort zone of pedestrians at current time-step.    

Inspired by \cite{liu2022socially}, we also design a prediction-based reward function called K-step lookahead reward to induce multiple robots to learn more social reasonable collaborative motion strategy. As represented in Fig.\ref{Fig1_Content}. The K-step lookahead reward configuration between the robot $i$ and the human $j$ is as follows: 
\begin{equation}\label{eq_kstep-reward}
	r_{K_{\text{step}}}^{R_iP_j} = \left\{\begin{array}{ll}
		\min_{tp=\{t+1,...,t+K\}}-e^{-k}  & \text{if} \min{\textbf{d}_{R_i^\prime P_i^\prime}^{tp}} < d_{\text{comfort}} \\
		0 & \text{otherwise}
	\end{array}\right.
\end{equation}  
$t$ is the current time-step. $d_{\text{comfort}}=0.3$ is the comfort threshold of social distance. We consider that during the process of the human-robot interaction, the intrusion action of robots can cause  "discomfort" and deliberate avoidance behavior of pedestrians.  ${\textbf{d}_{R_i^\prime P^\prime}^{tp} = [d_{R_i^\prime P_j^\prime}^{t+1}-r_{R_i}-r_{P_j},...]}$ represents the lookahead social distance between the robot $i$ and the pedestrian $j$ in its FOV. The final K-step lookahead reward of the robot $i$ is the minimum of all $r_{K_{\text{step}}}^{R_iP_j}$ where $j=\{1,2,...,m\}$ represents the index of the pedestrian that appears in FOV of the robot $i$. 
\begin{equation}\label{eq_kstep-reward_tot}
	r_{K_{\text{step}}} = \min_{j=\{1,2,...,m\}} r_{K_{\text{step}}}^{R_iP_j}
\end{equation} 

In addition, it is particularly noteworthy that the implementation of the K-step lookahead reward setting relies on two key points. First, we need a pre-trained trajectory prediction module (e.g., Social-GAN \cite{Gupta_2018_CVPR}, MID \cite{Gu_2022_CVPR}, etc.) to predict the K-timestep trajectory evolution of pedestrians in FOV at each timestep in the lookahead virtual time domain. Second, for the motion evolution of the robot $i$ in the lookahead virtual time domain, we choose to perform a uniform robot state update according to the current evaluated real action pair. This trick accelerates the training process, and allows to focus on quantifying the social reasonableness of the policy at the real timestep of the moment. 

To sum up, our final joint reward setting of multi-robot social-aware cooperative planning task is as follows:
\begin{equation}\label{eq_joint_rew}
	r_{\bm{R}} = \sum_{i=1}^{N} r_{R_i}
\end{equation} 
where $N$ is the number of robots in the environment.This mode of joint reward decomposition specifies the contribution of each agent to the team, and effectively attenuates the credit assignment issue in MARL setting. 

\subsection{Algorithm Description of MSA3C}
The overall MSA3C algorithm architecture of our multi-robot social-aware cooperative planning method is shown in Fig. \ref{Fig2_Tot_Alg_Arch}. MSA3C belongs to the CTDE paradigm. The CTDE paradigm-based cooperative planning inherits the advantages of centralized methods and decentralized methods respectively \cite{9773295}. We design the local TSG-based social encoder module to perform the social-interaction hidden state extraction, and handle the problem of variable input dimension caused by the dynamic change of pedestrians in the limited FOV of each robot. By combining with the parameter-sharing mechanism, the scalability of the planner is effectively improved. Meanwhile, we utilize multi-head global observation attention module as an alternative to the traditinoal concatenation operation  \cite{9773295, 9750312, wang2020r}. This approach allows for better aggregation of sensing information shared between robots, which can be leveraged to better guide policy network updates.  

\subsubsection{Rollout Replay Buffer}  
\begin{figure}[!t]
	\centering
	\includegraphics[width=3.3 in]{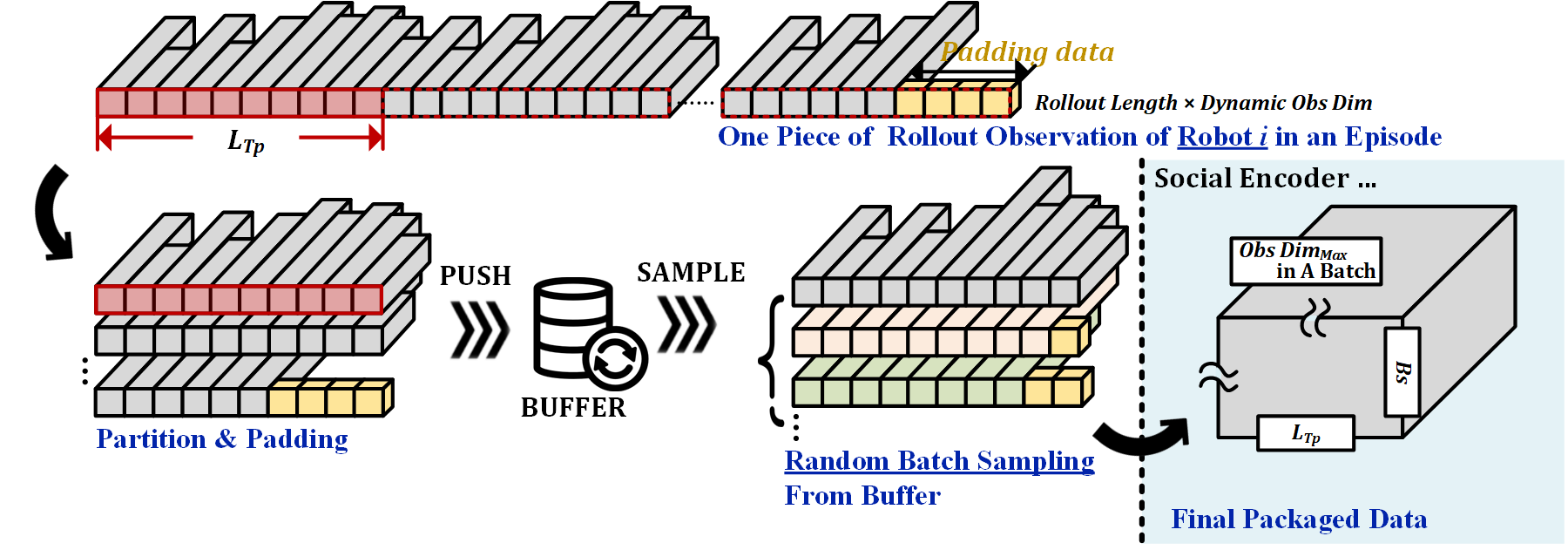}
	\caption{Rollout data processing pipeline: from rollout replaybuffer to the social encoder.}
	\label{Fig3} 
\end{figure} 

In the training phase, we need to push the joint interaction experience generated by all robots to the replaybuffer. Different from previous approach of adding transitions \cite{9773295}, these data need to be partitioned and stored in a fixed timestep sequence owing to the utilization of GRU-based local TSG in the later module. As shown in Fig. \ref{Fig3}, we design the rollout replaybuffer to deal with the rollout data and record padding positions used to align the timestep length in order to mask those pseudo-data in the subsequent loss back-propagation process. 

\subsubsection{Social Encoder}   
As shown in Fig. \ref{Fig3}. Although we align the timestep length of each batch,  the observation dimension is inconsistent due to the dynamic changing of the number of pedestrians per timestep in the limited FOV of each robot. So, the first step is to align the dimension of maximum observation dimension in each batch via the data packaging module in Fig. \ref{Fig2_Tot_Alg_Arch}. Then, the encapsulated data are fed into the local TSG as the input of spatial-edge RNN, temporal-edge RNN and node RNN, respectively.  

The TSG is widely applied in the field of pedestrian trajectory prediction \cite{8460504, 8794134, Huang_2019_ICCV}. In this paper, we utilize part of this graph to extract the human-robot interaction feature, and take this social feature as input to the MARL module. The social feature contains historical information and can reflect the potential importance of each human in FOV for the robot to make the next decision. The spatial-edge RNN of the TSG is responsible for capturing the spatial information such as relative orientation and distance of each robot with respect to other agents (e.g., humans, other robots, etc.) in its limited FOV. The specific procedure is as follows: 
\begin{equation}\label{SpatialRNN}
h_{r_i\textbf{A}}^t= GRU(h_{r_i\textbf{A}}^{t-1}, \phi(\textbf{x}^t_{r_i\textbf{A}};W_{\text{spatial}}^{\text{emb}});W_{\text{spatial}}^\text{h})
\end{equation} 
where $\textbf{x}^t_{r_i\textbf{A}}$ represents the relative position vector between robot $i$ and all other agents in its FOV at timestep $t$. $\phi$ is the embedding layer, including a fully connect (FC) layer and a layer normalization (LN) layer. $W_{\text{spatial}}^{\text{emb}}$ are the embeding weights. $h_{r_i\textbf{A}}^t$ is the hidden state of the GRU at timestep $t$. ${W_{\text{spatial}}^\text{h}}$ are the weights of spatial-edge RNN. 

Temporal-edge RNN is responsible for capturing the position evolution representation of current robot in adjacent frames. The specific process is similar to (\ref{SpatialRNN}): 
\begin{equation}\label{TempRNN}
h_{r_ir_i}^t= GRU(h_{r_ir_i}^{t-1}, \phi(\textbf{x}^t_{r_ir_i};W_{\text{temporal}}^{\text{emb}});W_{\text{temporal}}^{\text{h}})
\end{equation} 
where $\textbf{x}^t_{r_ir_i}$ is the position changing vector of the robot $i$ at adjacent timesteps. Then, we put $h_{r_ir_i}^t$ and $h_{r_i\textbf{A}}$ over a dot product multi-head social attention module to obtain the attention scores of different human-robot interactions. The specific process is shown as follows:
\begin{equation}\label{Attn}
\textbf{x}_{\text{attn}_{r_i}}^t=   \text{softmax}(F_\text{scale}W_q h_{r_i\textbf{A}}^t {h_{{r_ir_i}}^{tT}} W_k^T) \cdot (W_v h_{r_i\textbf{A}}) 
\end{equation} 
where $F_{scale}$ is the scale constant. $W_q$ are the Query weights, $W_k$ are the Key weights, and $W_v$ are the Value weights. 
Finally, we concatenate $\textbf{x}_{\text{attn}_{r_i}}$ with ego state of robot $i$, and send them to the node RNN after embedding operation. The embedding feature is passed through the final feature extraction layer to generate the fixed-length social feature: 
\begin{equation}\label{SocialFeatureEq1}
h_{r_i}^t= GRU(h_{r_i}^{t-1}, \phi(\text{cat}[\textbf{x}_{{\text{attn}}_{r_i}}^t,\textbf{x}_{r_i}^t];W_\text{node}^\text{emb});W_\text{node}^\text{h})
\end{equation} 
\begin{equation}\label{SocialFeatureEq2}
	\textbf{x}_{\text{social}_i}^t = \phi(h_{r_i}^t; W_\text{social}^\text{emb})
\end{equation} 
$\textbf{x}_{\text{social}_i}^t$ would be fed to the MARL part in the later stage.

\subsubsection{Multi-robot Attention-based Actor Critic}  
 
Our MARL algorithm is on the basis of multi-agent local-and-global attention actor-critic (MLGA2C) in our previous work \cite{9953939}. Here, we replace local attention module in MLGA2C with social encoder to better help robot understand human-robot interaction in FOV to improve the safety of planning in human crowds. 

The final MSA3C integrates an global-attention-based critic network and a policy network.  Different homogeneous robots are parameter-sharing to improve the training efficiency. The purpose of global attention computation between robots in centralized value network is to capture the most important shared inter-robot feature for the current robot. This information aggregation pattern facilitates more accurate decision making by the current robot. In \cite{9773295}, we also empirically demonstrate its superioriy compare to other baselines in multi-robot cooperative planning tasks. Specifically, for the robot $i$, the joint action state evaluation Q can be described as follows: 
\begin{equation}\label{MLGA2C_TOTQ} 
\begin{aligned}
Q_{r_i}&=\phi_\text{emb}(\left[\text{softmax}(W_{q}^\text{RL}\phi_\text{ego}([\textbf{x}_{\text{social}_i},\bm{a}_i])(W_{k}^\text{RL}\phi_\text{others}^{-i})^T)) \cdot \right.  \\
	     & \left.   \text{\space\space\space}(W_v^\text{RL}\phi_\text{ego}),\phi_\text{ego} \right]) 
\end{aligned}
\end{equation}  
where $W_{q}^\text{RL},W_{k}^\text{RL},W_v^\text{RL}$ represent the weights of the dot product attention module between robots. We concatenate $\textbf{x}_{\text{social}_i}$ with corresponding batch of action data $\bm{a}_i$, and feed them into the embedding layer $\phi_\text{ego}$ of the ego-robot. $\phi_\text{others}^{-i}$ represents the embedding feature about social features and actions of all other robots. Then, we concatenate the inter-robot attention weights with $\phi_\text{ego}$ and feed them into multi FC layers. The output value of the final layer is the final centralized Q value of the robot $i$. The loss function of our centralized Q is defined as follows: 
\begin{equation}
	\begin{aligned}L_{Q}(\psi) &=\sum_{i=1}^{N} \mathbb{E}_{(\boldsymbol{o}, \boldsymbol{a}, \boldsymbol{r}, \boldsymbol{o}\prime, \bm{h}_\text{temp},\bm{h}_\text{spa}, \bm{h}_\text{node}) \sim D}\left[\sum_{j=\{1,2\}}\left(Q_{r_i}^{\psi_{j}}-y_{r_i}\right)^{2} \right]\end{aligned}
\end{equation}
where 
\begin{equation}
	\begin{aligned}
		y_{r_i}=&\mathbb{E}_{\boldsymbol{a}^{\prime} \sim \pi_{\theta}\left(\textbf{x}^\prime_{\text{social}_i} \right)}\left[\bm{r}+\gamma\left(-\alpha \log \left(\pi_{\theta}\left(\boldsymbol{a}_{i}^{\prime} \mid \textbf{x}^\prime_{\text{social}_i}\right)\right)\right.\right. \\
		&\left.\left.+\min \left(Q_{r_i}^{\bar{\psi}_{1}}, Q_{r_i}^{\bar{\psi}_{2}}\right)\right)\right].
	\end{aligned}
\end{equation}
$D$ represents our rollout replaybuffer. To avoid overestimation problem of classical deep Q learning, we introduce two separated Q heads $Q_{r_i}^{\psi_{1}}, Q_{r_i}^{\psi_{2}}$ and corresponding target Q heads $Q_{r_i}^{\bar{\psi}_{1}}, Q_{r_i}^{\bar{\psi}_{2}}$. $\pi_{\theta}$ is the policy network with shared parameters. $y_{r_i}$ is the soft TD target value. We have the same policy entropy setting about $y_{r_i}$ as in the previous paper \cite{9773295}. $\gamma$ is the discount factor. $\alpha$ is a trainable parameter that can be used to auto-adjust the exploration degree of the robot during the training phase.

The update of the policy network is consistent with our previous work \cite{9953939}: 
\begin{equation}
	\begin{aligned}\nabla_{\theta} J\left(\pi_{\theta}\right)=& \sum_{i=1}^{N} \mathbb{E}_{(\boldsymbol{o}, \boldsymbol{a},\bm{h}_\text{temp},\bm{h}_\text{spa}, \bm{h}_\text{node}) \sim D, \boldsymbol{a}_{i} \sim \pi_{\theta}}\left[\nabla_{\theta} \log \pi_{\theta}\right.\\&\left.\left(-\alpha \log\pi_{\theta}+\min \left(Q_{r_i}^{\psi_{1}}, Q_{r_i}^{\psi_{2}}\right)\right)\right]\end{aligned}
\end{equation}
Also, the reparameterization trick are deployed here to ensure the smooth back-propagation process: 
\begin{equation}
\boldsymbol{a}_{i} \sim \pi_{\theta}=\tanh \left(\mu_{\theta}+\sigma_{\theta} \odot \varepsilon\right) \text{ , } \varepsilon \sim \mathcal{N}(0,I)
\end{equation}
where $\mu_{\theta}$ and $\sigma_{\theta}$ are the outputs of policy parameters. 

\subsubsection{Decentralized Cooperative Planning} 
The decentralized execution of our MSA3C-based cooperative planning methods is much easier. As shown in Fig. \ref{Fig2_Tot_Alg_Arch},  each robot utilizes its own local observation to make decisions, and this process can be online and real-time. In the realistic deployment stage, this cooperative planner can be combined with the model predictive controller or others to achieve autonomous and collaborative operation of multiple robots.

\begin{table}[] 
	\centering
	\caption{The Basic Parameter Setting of the Environment and MSA3C }
	\label{Table_Exp_Setting}
	\begin{tabular}{lllc}
		\toprule[1pt] 
		\textbf{Env Setting}&\textbf{Value}&\textbf{RL Setting}&\textbf{Value}  \\  
		
		\hline 
		$r_\text{Safety}^{P}$&  0.5$\sim$1.3m     & $K_{\text{lookahead}}$ & 5                       \\  
		$r_\text{Safety}^{R}$ 		& 0.6m  & $r_{\text{time}}$ & -1e-3 \\
		$v_{\text{pref}}^P$  &    0.5$\sim$1.5m/s & $lr$ & 5e-4             \\ 
		$v_{\text{pref}}^R$  &    1m/s  & $\tau$ & 0.01                \\ 
		$d_{\text{comfort}}$  & 0.25m  & Policy delay & 2                          \\ 
		$R_{\text{scenario}}$ & [6m,8m,10m] & Batch size& 256 \\ 
		$N_{\text{peds}}$ & 5 $\sim$ 20 & $\alpha_\text{init}$ & 0.02 \\  
		$N_{\text{robots}}$ & 3  & Buffer size & 2e5\\ 
		FOV & 2$\pi$, [5m,10m] & episode & 5e4\\   
		Timestep & 0.25s  & $l_{\text{rollout-tps}}$ & 10\\
		\bottomrule[1pt]  
	\end{tabular} 
\end{table} 


\section{Experiments and Discussion}  

\subsection{Experimental Configuration} 
\subsubsection{Basic Setting} 
To verify the performance of MSA3C, we have developed a Python-based gym simulation platform for multi-robot cooperative planning in pedestrians on the basis of \cite{8593871,8794134}. As shown in Fig. \ref{Fig1_Content}, we utilize the circle to represent the safety zone and the social comfort zone of pedestrians. At the beginning of each episode, we randomly assign the safety radius, the social comfort radius, and the preferred velocity to simulate different number, body shape, and motion state of each pedestrian unit. We assign each pedestrian a random goal change probability parameter with the value range $[0.2,0.3]$ to simulate the motion uncertainty of humans during the training phase.  Also, we add timestep-varying random noise to the safety radius of each pedestrian unit to simualte the sensing error of the robot. Different from previous work \cite{8794134}, we utilize social-force (SF) model to simulate the movement and interaction of pedestrians. SF model is able to describe the self-organization of several group effects of the observed pedestrian behavior more realistically \cite{mehran2009abnormal}. In addition, we adopt a non-collaborative human-robot interaction mode. Pedestrians are defaulted to be the influencer, robot are the reactor. This setting prevents robots from learning aggressive cooperative planning policy to obtain high returns \cite{liu2022socially}. It is worth mentioning that we adopt a multi-stage training scheme.  In stage I, our goal is to encourage multiple robots to learn a high efficiency cooperative pattern. In stage II, we introduce the K-step prediction reward setting to develop the comfort social awareness of each robot, while employing a larger collision penalty (-5) to ensure the planning safety. 

All experiments are performed on a server with an Intel(R) Xeon(R) Silver 4214CPU and a GeForce RTX 3090 GPU. We have summarized the basic parameter setting in Table \ref{Table_Exp_Setting}.

\subsubsection{Network Setting of MSA3C} 
MSA3C consists of the social encoder, the attention-based double value net, and the policy net. The social encoder including the temporal-edge RNN with the size of [2,64,256], the spatial-edge RNN with the unit dimension of [2,64,256], the multi-head social attention block with three embedding layers ($W_q, W_k, W_v$) of dimension [256,128], and the node RNN with the unit dimension [7,64,128,256]. Each RNN module is equipped with the LayerNorm layer and the LeakyReLU activation function. The global attention-based double value net contains two critic heads. Each critic head is composed of the ego-embedding layer with the size of [258, 128], the others-embedding layer with the size of [258, 128], the multi-head global attention block with three embedding layers ($W_q^{\text{RL}},W_k^{\text{RL}}, W_v^\text{RL}$) of dimension [128, 128], and the Q value output block with the unit dimension [256, 128, 128, 1]. Similarly, each sub-block of the global critic head is equipped with the Layernorm layer and the LeakyReLU activation function.  The decentralized policy network contain three hidden fully connected layers with the dimension of [256,128,128,128,2].  

\subsection{Metrics}
To quantify the performance of different algorithms, we set the following evaluation metrics:
\begin{enumerate}[]
	\item ``CSR": Success rate of co-planning process. We specify that all robots reach their target points within the limited timesteps to be considered as a "success". 
	
	\item ``CR": Collision rate. CR describes the rate that the robot collides with other agents. 
	
	\item ``APL": Average co-planning path length of all robots. This metric describes the cooperative capability of different algorithms. 
	
	\item ``NTC": N-robot co-planning timestep comsumption. This metric describes the cooperative efficiency of different algorithms
	
	\item ``CIR": Comfort zone intrusion rate. This metric reflects the social awareness of robots during the co-planning process.
\end{enumerate} 

\begin{table}[] 
	\label{QuanExpTab}
	\caption{Summary table of quantitative experiment results in different pedestrian participation scenarios with limited timesteps (150)}
	\resizebox{\columnwidth}{!}{
		\begin{tabular}{@{}lclllll@{}}
			\toprule 
			\multirow{2}{*}{\textbf{Algorithms}} &
			\multirow{2}{*}{\textbf{Scene}} &
			\multicolumn{5}{c}{\textbf{Different Metrics}} \\ \cmidrule(l){3-7} 
			\multicolumn{1}{c}{} &
			&
			\multicolumn{1}{c}{\textbf{CSR\%$\uparrow$}} &
			\multicolumn{1}{c}{\textbf{CR\%$\downarrow$}} &
			\multicolumn{1}{c}{\textbf{APL$\downarrow$}} &
			\textbf{NTC$\downarrow$} &
			\multicolumn{1}{c}{\textbf{CIR\%$\downarrow$}} \\ \midrule
			MASAC-F10                      & \textbf{5p3r}  & Nan & Nan & Nan & Nan & Nan  \\ \midrule
			MLGA2C-F10                     & \textbf{5p3r}  & 98.0    & 44.2    &   \textbf{26.8}  & \textbf{45.7} & 10.2 \\ \midrule
			\multirow{3}{*}{\textbf{MSA3C-F10}}  & \textbf{5p3r}  &  92.6   &  0.9   &  \textbf{27.7}   &  \textbf{51.4}   &  3.0     \\ \cmidrule(l){2-7} 
			& 10p3r &  80.2  &  2.2 &   31.8  &  71.6 &   4.1   \\ \cmidrule(l){2-7} 
			& 20p3r &  61.4   &  5.8   &   36.9  &  81.4  &  6.7   \\ \midrule
			\multirow{3}{*}{\textbf{MSA3CPred-F10}} & \textbf{5p3r}  &  \textbf{98.8}   &  2.4   &  \textbf{28.7}   &  \textbf{51.7}   &  \textbf{1.2}     \\ \cmidrule(l){2-7} 
			& 10p3r & \textbf{99.8}  &   2.5 &  33.1 &  65.2  & \textbf{1.6}   \\ \cmidrule(l){2-7} 
			& 20p3r & \textbf{92.6} & 8.4 &  41.2  &  81.0 &  \textbf{2.9} \\ \midrule
			\multirow{3}{*}{\textbf{MSA3CPred-F5}}  & \textbf{5p3r}  &  99.2  &  11.6  & 34.2  & 65.8  & 2.3     \\ \cmidrule(l){2-7} 
			& 10p3r &  97.2   & 14.9  &  40.2 &  79.3   &  2.7    \\ \cmidrule(l){2-7} 
			& 20p3r &  89.5  &  16.6   &  43.6  &  84.3   &  3.8     \\ \midrule
			*SARL-F10   & \textbf{5p3r}  &  85.2   &  12.2   &  \textbf{40.4}   & \textbf{56.5}   &  4.9      \\ \midrule
			\multirow{3}{*}{*ORCA-PS-F10}     & \textbf{5p3r}  &  98.4   &  2.8   &  30.2   &   66.4  &  1.4     \\ \cmidrule(l){2-7} 
			& 10p3r &  92.4   &  3.5  &  38.2   &  89.7   &  1.6       \\ \cmidrule(l){2-7} 
			& 20p3r &  60.4   &  7.8  &  39.5   &  102.9  & 2.5  \\ \midrule
			\multirow{3}{*}{*SF-F10}       & \textbf{5p3r}  & 98.0  &  35.5  & 33.1  & 80.8    &  7.3    \\ \cmidrule(l){2-7} 
			& 10p3r & 68.2 &  77.9   &  37.1  &  89.4   &  9.6   \\ \cmidrule(l){2-7} 
			& 20p3r &  71.2   &   135.1  &  42.9   &  99.0   &   12.5  \\ \bottomrule
		\end{tabular}
	}
\end{table}


\subsection{Quantitative \& qualitative analysis} 
\begin{figure*}[!t]
	\centering
	\label{}
	\includegraphics[width=6.2 in]{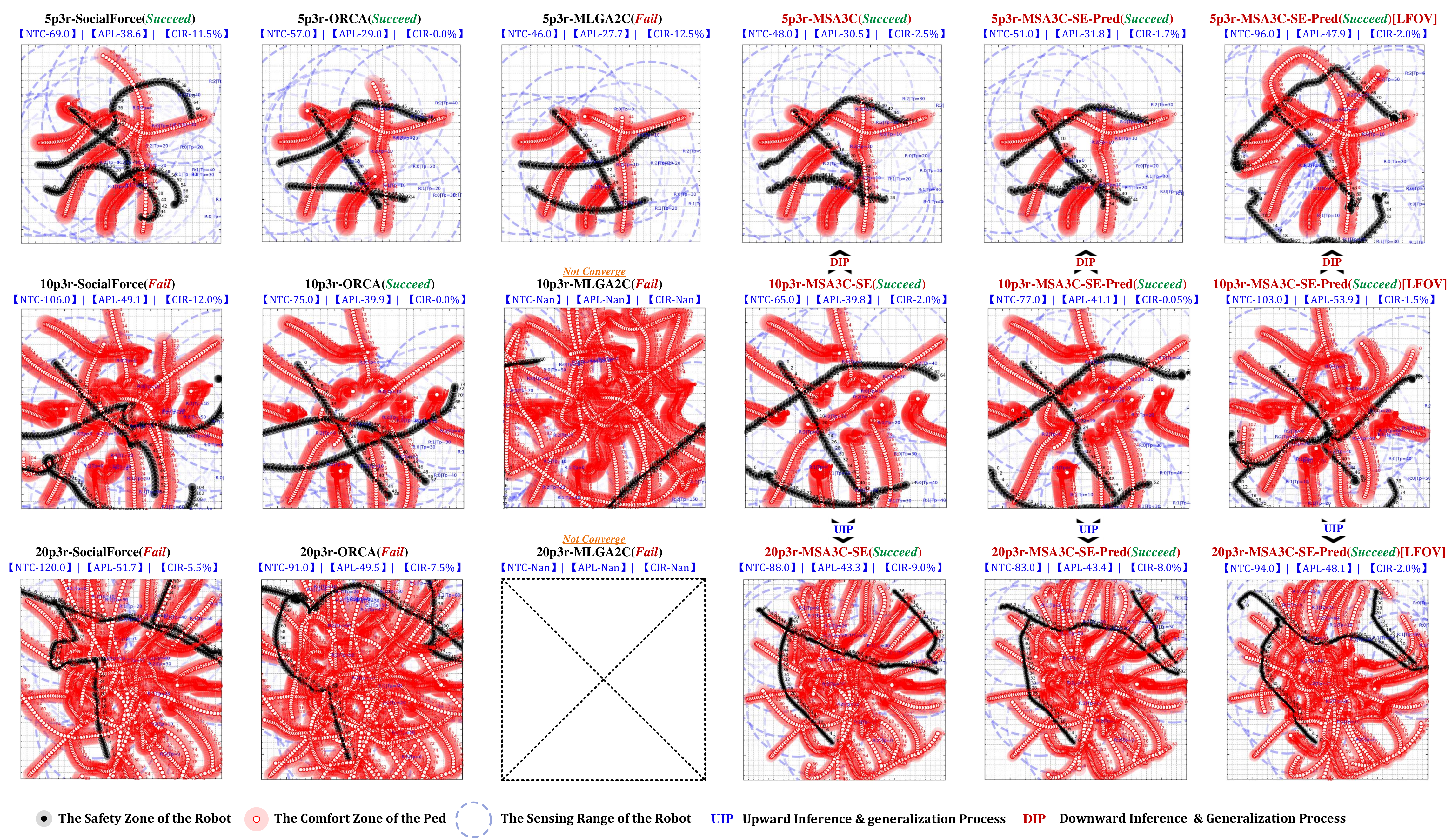}
	\caption{Collaborative trajectories of multiple robots generated by different co-planning algorithms in a pedestrian interaction scenario with fixed random seeds. The red curve represents the predestrian motion trajectory, and the black curve represents the robot motion trajectory.}
	\label{Fig4_CoTrajsPlotFig} 
\end{figure*}  

\begin{figure}[!t]
	\centering 
	\includegraphics[width=2.8 in]{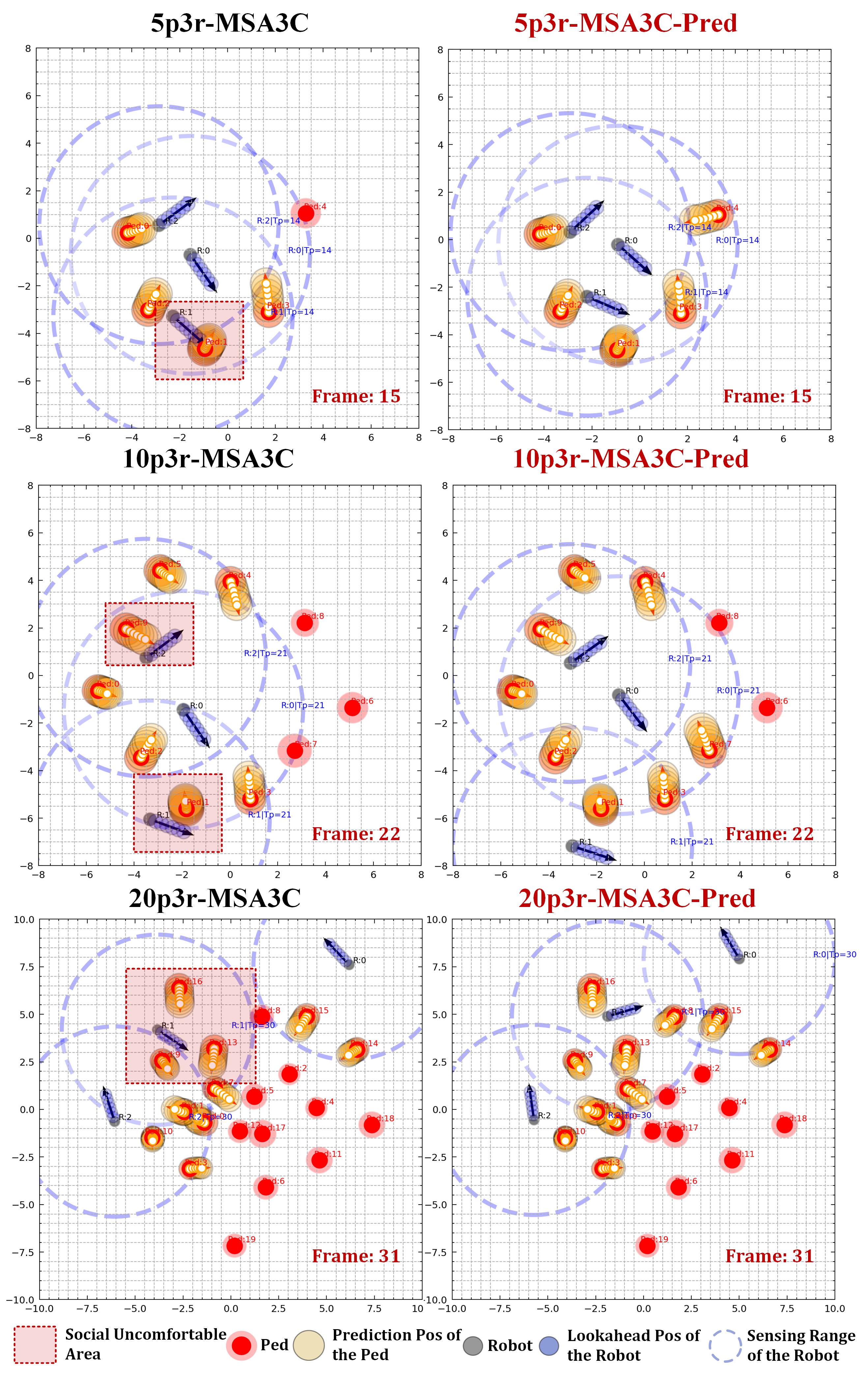}
	\caption{The social relationship plots and the visualization of K-step prediction of human comfort zones for MSA3C and full version MSA3CPred at specific frame extracted from Fig. \ref{Fig4_CoTrajsPlotFig}}
	\label{Fig5_FramePlotFig} 
\end{figure} 

Table \ref{QuanExpTab} summarizes the performance of different algorithms. "NpMr" represents that there are "N" pedestrians and M robots in current scenario. "-FX" represents the current FOV of each robot is "X"m. For each "NpMr" scenario, we randomly run 500 times against different algorithms. Each algorithm is run with the same random seed to ensure the consistency of the pedestrian randomness and the same initial positions of robots. Most importantly, we specify 150 limited timesteps for each scenario to compare the efficiency of different algorithms.

\subsubsection{Baselines} 
We set up multiple baseline algorithms for comparison. We first select off-policy MASAC under local observation condition as the MARL architecture. MASAC-based co-planning method has achieved good collaborative performance in pure robot environment with the naive concatenation global information aggregation pattern \cite{9354210}. SARL in \cite{zhou2021r} is a representative decentralized social aware planning method with the single agent RL (SARL) setting. We utilize this method to verify the effect of offline multi-robot information aggregation on improving the overall co-planning performance. ORCA-PS-F10 stands for the typical optimal reactive-based multi-robot co-planning method under local perfect sensing condition in \cite{berg2011reciprocal}. SF stands for the classical social force approach \cite{mehran2009abnormal}. 
\subsubsection{Ablation Setting}
To verify the performance of each module in our method, we also couple various configurations in our quantitative experiments. MLGA2C is our previous work that achieves SOTA cooperative performance in dense pure robot scenarios \cite{9953939}. In MSA3C setting in Table \ref{QuanExpTab}, we replace the attention-based local observation processing head with the TSG-based parameter-sharing social encoder and introduce the rollout data processing pipeline. MSA3CPred is the full version method we describe above. It should be noticed that all of these ablation methods are trained in 10p3r scenario and then directly inferred in 5p3r as well as dense 20p3r scenarios without any parameter adjustment. 
\subsubsection{Quantitative Analysis}  
First, we would analyze the basic 5p3r group. Since MASAC cannot converge in this pedestrian participation environment setting, the detailed data on each metrics cannot be output. A comparison on APL and NTC metrics show that our previous global attention critic-based MLGA2C still obtains excellent cooperative performance \cite{9354210}. However, the simple proximity function cannot inspire the robot to learn to correctly handle social relationships with uncertain pedestrians in its FOV. Higher CR\% and CIR\% imply that the robot develops an aggressive policy with the primary target of goal reaching. As for SARL-F10, we adopt a recent decentralized social aware planning method in \cite{zhou2021r}. Although SSR\% and CIR\% performance of SARL-F10 is not bad, the lack of global information makes it cannot achieve good performance in cooperative metrics, like APL and NTC. Our MSA3Cs under different settings not only both have great CSR\% and lower CR\% compared to other baselines, but also produce great cooperative performance on the metrics of APL and NTC between multiple robots. Most importantly, with the help of the TSG-based social encoder and the K-step lookahead reward function, our MSA3Cs help robots generate more reasonable social awareness which leads to lower CR\% and CIR\%. 

In the more complex and dense 10p3r and 20p3r scenarios, basic MLGA2C fails to converge. With the introduction of our TSG-based social encoder, MSA3C-F10 can aid robots to make more socially safe decisions. After further coupling K-step lookahead reward function, MSA3CPred-F10 preforms well in different metrics especially in CIR\%. This means robots under this setting generate social comfort awareness. We also find that the K-step lookahead trick motivates robots to master the skill of avoiding dense social interaction regions. This skill avoid freezing robot issue, and help full version of MSA3CPred achieve higher CSR\% in limited timesteps in comparison to other MARL-based methods and reactive-based methods.  

Furthermore, we narrow the sensing range of each robot to 5m. The results of MSA3CPred-F5 show that the global attention module-based MARL architecture can still effectively aggregate the limited local observation of each robot and help multiple robots obtain decent collaborative performance on CSR\%, APL, and NTC. The TSG-based social encoder and K-step lookahead trick help robots to maintain efficient co-planning ability while still achieving a good level of social safety and social comfort on CR\% and CIR\% metrics in human-robot interacting process.

\subsubsection{Qualitative Analysis} 
The co-planning trajectories of robots driven by different algorithms in a random scenario are shown in Fig. \ref{Fig4_CoTrajsPlotFig}. Meanwhile, we extract one critical frame from this co-planning process to visualize the effect of the pedestrian trajectory prediction module-based K-step lookahead reward function on human-robot social interaction process in Fig. \ref{Fig5_FramePlotFig}. It is necessary to state that these plots are generated with the same random seed. The denser the red trajectories of pedestrian, the more timesteps robots consume for co-planning process. First, we can find that the MLGA2C group generates co-planning trajectories with better smoothness and straightness, but fails to complete these three tests without collision. Reactive-based methods are weak in generalizing across different scenarios and have inconsistent performance. All of the other three MSA3C-based settings help robots complete co-planning process without the human-robot collision. Further study of the trajectories reveals that MSA3CPred can help the robot adjust its policy to avoid dense social interaction regions during the planning process. Fig. \ref{Fig5_FramePlotFig} can illustrate this conclusion more visually. We can find that the full version MSA3CPred helps robots produce more farsighted policies in the three scenarios 5p3r, 10p3r, and 20p3r. These motion policies not only guarantees the social safty of robots in the present timestep, but also ensures robots avoid entering the K-step prediction comfort zone of pedestrians in the future prediction time domain. This mechanism also effectively motivates robots escape from dense social interaction areas that are prone to freezing issues. These qualitative analyses also coincides with the conclusions drawn in quantitative experiments.

\section{Conclusion} 
We present a brand new social aware multi-robot cooperative planning method MSA3C on the basis of MARL architecture with attention mechanism.  The algorithm generally follows the CTDE paradigm and introduce the multi-head attention-based centralized critic network to better aggregates the local information from each robot.  By replacing the simple proximity function with TSG-based social-encoder, our model allows each robot to better understand the social importance of each pedestrian in its FOV.  Also, we introduce the K-step lookahead reward setting to alleviate the intrusion of the robot into the social comfort zone of pedestrians, and avoid short-sighted decision making process of each robot.  Through quantitative and qualitative analyses in multiple pedestrian participating co-planning scenarios, we show our MSA3C outperforms multiple baselines.

\bibliographystyle{IEEEtran}
\bibliography{Reference.bib}

\newpage

 

%
%


\end{document}